\documentclass[american]{article}
\usepackage[T1]{fontenc}
\usepackage{textcomp}
\usepackage[utf8]{inputenc}
\usepackage{refstyle}
\usepackage{booktabs}
\usepackage{graphicx}
\usepackage{natbib}

\makeatletter

\AtBeginDocument{\providecommand\Figref[1]{\ref{Fig:#1}}}
\AtBeginDocument{\providecommand\Eqref[1]{\ref{Eq:#1}}}
\AtBeginDocument{\providecommand\Tabref[1]{\ref{Tab:#1}}}
\providecommand{\tabularnewline}{\\}
\RS@ifundefined{subsecref}
  {\newref{subsec}{name = \RSsectxt}}
  {}
\RS@ifundefined{thmref}
  {\def\RSthmtxt{theorem~}\newref{thm}{name = \RSthmtxt}}
  {}
\RS@ifundefined{lemref}
  {\def\RSlemtxt{lemma~}\newref{lem}{name = \RSlemtxt}}
  {}

\usepackage{arxiv}
\usepackage{csquotes}

\makeatother

\usepackage{babel}

\begin{document}
\title{Task-Driven Fixation Network: An Efficient Architecture with Fixation
Selection}
\author{Shuguang Wang, Yuanjing Wang}
\maketitle
\begin{abstract}
This paper presents a novel neural network architecture featuring
automatic fixation point selection, designed to efficiently address
complex tasks with reduced network size and computational overhead.
The proposed model consists of: a low-resolution channel that captures
low-resolution global features from input images; a high-resolution
channel that sequentially extracts localized high-resolution features;
and a hybrid encoding module that integrates the features from both
channels. A defining characteristic of the hybrid encoding module
is the inclusion of a fixation point generator, which dynamically
produces fixation points, enabling the high-resolution channel to
focus on regions of interest. The fixation points are generated in
a task-driven manner, enabling the automatic selection of regions
of interest. This approach avoids exhaustive high-resolution analysis
of the entire image, maintaining task performance and computational
efficiency.
\end{abstract}

\section{Introduction}

The human brain receives a vast amount of input information at every
moment. However, most of this information is unrelated to the current
task and should be filtered out to focus on the important information.
On the other hand, the brain's information processing capacity is
limited, allowing it to handle only critical and essential information
while ignoring less significant inputs.

Taking the human visual system as an example, the visual system has
high resolution only within a very narrow 2° range at the center of
the visual field and relatively good resolution within a 5° range.
In contrast, the peripheral regions of the visual field have very
low resolution. Assuming an observation distance of 30 cm (typical
for reading), the high-resolution area is only about 1 cm wide. Therefore,
the visual system uses eye movements to continuously shift the fixation
point, focusing on regions of interest in the visual field in a serial
manner. This strategy effectively replaces high-resolution analysis
of the entire visual field.

Psychologists have conducted extensive research on eye movements and
fixation mechanisms. For example, Yarbus (\cite{yarbusEyeMovementsVision1967})
demonstrated that fixation is an indispensable component of the visual
system. Even when viewing static images, the fixation point continuously
moves to focus on important parts of the image. The selection of fixation
points is task-related. In the same scene, different tasks lead to
significantly different distributions and trajectories of fixation
points.

Research on eye movements during reading reveals that during silent
reading, the mean saccade distance is approximately 8 characters or
two degrees (\cite{raynerEyeMovementsReading1998}). This may implies
that each fixation provides effective input for around 8 characters.
This differs significantly from modern neural networks, which often
handle input widths of 32K tokens or more.

There are significant differences in eye movements and fixations between
children and adults, with preschool children exhibiting more frequent
small saccades, less stable fixation, longer saccadic latency, and
poorer localization accuracy (\cite{raynerEyeMovementsReading1998}).
This suggests that the development of a stable and accurate eye movement
and fixation mechanism may require prolonged learning through experience
rather than reliance on predefined rules.

Some studies have also revealed that fixation helps simplify complex
tasks by breaking them down into object-centered, sequential sub-tasks.
Additionally, humans tend to prefer serial fixations, which reduce
the load on short-term memory, rather than operating at its maximum
capacity (\cite{ballardMemoryRepresentationsNatural1995}).

Treisman’s (\cite{treismanFeatureintegrationTheoryAttention1980})
Feature Integration Theory suggests that visual information is first
processed in parallel across a series of independent feature dimensions,
such as color, orientation, spatial frequency, brightness, and motion
direction. These features are then integrated through serial focused
attention on specific locations to form perceptions of target objects
defined by conjunctions of features. Without focal attention, features
across dimensions may remain floating, potentially forming illusory
combinations.

Based on Feature Integration Theory, a visual model for salient region
detection was proposed (\cite{kochShiftsSelectiveVisual1985,ittiModelSaliencybasedVisual1998}).
This bottom-up model relies on predefined rules, where salient regions
are defined as areas that significantly differ from their surroundings
in specific feature dimensions. However, this approach does not account
for the specificity of particular visual tasks and is therefore more
suited for general-purpose scene analysis without defined objectives.

In fact, eye movements and fixations may be influenced by a combination
of factors, including saliency, target features, and scene context
(\cite{ehingerModellingSearchPeople2009}). Some models have attempted
to integrate bottom-up and top-down information to determine fixation
points (\cite{olivaTopdownControlVisual2003,petersBottomupIncorporatingTaskdependent2007,zhangSUNBayesianFramework2008,gaoDecisionTheoreticSaliencyComputational2009,yulinxieBayesianSaliencyLow2013}),
In this context, bottom-up refers to saliency maps derived from low-level
features, while top-down encompasses target features and scene context.
In recent years, a growing trend has been the use of deep neural networks
to detect salient regions or generate fixation points (\cite{vigLargeScaleOptimizationHierarchical2014,huangSALICONReducingSemantic2015,kummererDeepGazeBoosting2015,kummererDeepGazeIIReading2016,panSalGANVisualSaliency2018,wangInferringSalientObjects2020,wangRevisitingVideoSaliency2021,yanReviewVisualSaliency2021}).

To date, most of these models have primarily focused on simulating
human visual attention mechanisms, detecting salient regions, or producing
fixation point distributions and saccadic trajectories similar to
those of humans.

In this paper, we propose a novel model named Task-Driven Fixation
Network (TDFN). TDFN is designed to perform specific tasks (e.g.,
classification) efficiently, with fixation integrated as a submodule
to capture high-resolution information from the most task-relevant
regions. The model initially leverages global low-resolution information
to execute the primary task and dynamically selects fixation points
to sequentially incorporate high-resolution key regions as needed,
thereby enhancing task performance.

The primary objective of TDFN is to reduce the computational complexity
and scale of neural networks, enabling cost-effective task execution
without compromising performance metrics. Experiments on the MNIST
dataset validate the model's effectiveness, demonstrating its ability
to achieve high task performance with significantly reduced resource
requirements.

\section{Model}

\begin{figure}
\centering
\includegraphics[scale=0.65]{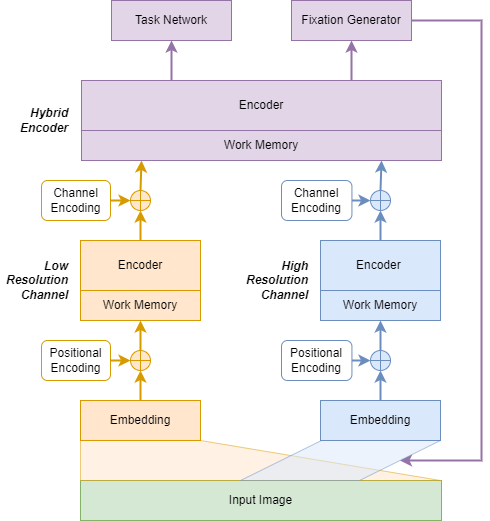}

\caption{TDFN architecture. }\label{fig:TDFN-architecture.}

\end{figure}

TDFN is implemented based on a Transformer architecture (\cite{vaswaniAttentionAllYou2017}),
as shown in \Figref{TDFN-architecture.}. Transformers are computationally
efficient, highly scalable, and have been successfully applied beyond
natural language processing, such as in ViT for image recognition
(\cite{dosovitskiyImageWorth16x162021}). A key advantage of Transformers
is their ability to capture global correlations through the multi-head
attention mechanism. However, computing these global correlations
introduces significant computational overhead. In TDFN, we leverage
the visual fixation mechanism to achieve complex tasks with a small-scale
network, mitigating this overhead.

We use image recognition as an example to illustrate this model.

As depicted in \Figref{TDFN-architecture.}, the main components of
TDFN include the Low-Resolution Channel (LRC), the High-Resolution
Channel (HRC), and the Hybrid Encoder (HE). These components share
a common structure comprising a Transformer encoder and a work memory.
The work memory temporarily stores the input token sequences of each
component and serves as an implicit structure of the Transformer,
enabling the parallelization of serial input sequences. Here, we explicitly
highlight the work memory to emphasize that the output of the HRC
is sequentially incorporated into the work memory of the HE.

The operational process of the model is outlined as follows:
\begin{enumerate}
\item The global low-resolution image is divided into patches, embedded,
and position-encoded before being fed into the work memory of the
Low-Resolution Channel (LRC).
\item The LRC encodes the token sequence in its work memory and outputs
a predefined set of tokens to the work memory of the Hybrid Encoder
(HE).
\item The HE encodes its token sequence and outputs task-relevant tokens
to the task network for task completion. Concurrently, it generates
a token for the fixation point generator to determine the next fixation
point.
\item Centered on the current fixation point, a high-resolution Region of
Interest (ROI) is cropped from the input image, divided into patches,
embedded, and position-encoded before being fed into the work memory
of the High-Resolution Channel (HRC).
\item The HRC encodes its token sequence and outputs a predefined set of
tokens, which are appended to the HE's work memory.
\item Steps 3 to 5 are repeated until the task performance metric meets
the desired threshold or further improvement through fixation becomes
infeasible.
\end{enumerate}

\subsection{Information Transfer Between Modules}

To reduce the network size, in our implementation, the LRC and HRC
only output two tokens: the class token (cls\_token) and the reconstruction
token (rec\_token). The cls\_token is used for classification, while
the rec\_token is used for image reconstruction and fixation point
generation.

As a result, the organization of the HE's work memory is as follows:

\[
memory^{he}=\left\{ ct^{lrc},rt^{lrc},ct_{1}^{hrc},rt_{1}^{hrc},ct_{2}^{hrc},rt_{2}^{hrc},\cdots\right\} 
\]

where $ct^{lrc}$ and $rt^{lrc}$ correspond to the class token (cls\_token)
and reconstruction token (rec\_token) derived from LRC. Similarly,
$ct_{i}^{hrc}$ and $rt_{i}^{hrc}$ are the class token and reconstruction
token derived from HRC, $i$ denotes the $i$-th fixation.

\subsection{Task Networks}

On top of the HE, we implement two task networks: a classifier and
an image reconstructor. Both the classifier and the image reconstructor
are simple two-layer fully connected feedforward networks. The output
layer of the classifier uses a softmax activation function.

The image reconstructor is designed to reconstruct the input image.
Its inclusion in the TDFN architecture serves to encourage a balance
between the extraction of categorical and structural features during
training. Both types of features are essential for generating effective
fixation points. The total loss function for classification training
is defined as follows:

\begin{equation}
TaskLoss=ClassLoss+\alpha\cdot ReconLoss\label{eq:taskloss_formula}
\end{equation}

Here:
\begin{itemize}
\item $ClassiLoss$ is the classification loss for the classifier, defined
using cross-entropy.
\item $ReconLoss$ is the reconstruction loss for the image reconstructor,
defined using mean squared error.
\item $\alpha$ is a tuning coefficient with a value range of {[}0, 1.0{]}.
\end{itemize}

\subsection{Fixation Point Generator}

The fixation point generator is a simple two-layer fully connected
feedforward network. The hidden layer uses the Leaky ReLU activation
function, while the output layer employs a softmax activation function.
The input to the fixation point generator is the rec\_token produced
by the HE, and its output is a saliency map where each point represents
its significance as a probability.

A Monte Carlo sampling method is applied to the saliency map to obtain
the fixation point.

\subsection{Training Strategy}

To simplify training, we adopt a step-wise approach.

First, the network is trained for the primary task using the loss
function defined in \Eqref{taskloss_formula}, with fixation points
generated randomly during this phase. The objective of task training
is to enable the model to learn robust feature representations.

After completing task training, the fixation points are optimized
using a reinforcement learning strategy. In this phase, all other
network parameters are frozen, and only the fixation point generator
is updated. For each fixation point $n$, the reward and loss function
for reinforcement learning are defined as follows:

\begin{equation}
Reward_{n}=TaskLoss_{n-1}-TaskLoss_{n}
\end{equation}

\begin{equation}
L_{n}=-Reward_{n}\cdot\log\left(p_{n}\right)
\end{equation}

Here:
\begin{itemize}
\item $p_{n}$ is the probability of generating the $n$-th fixation point.
\item $TaskLoss_{n-1}$ is the task loss before incorporating the $n$-th
fixation point.
\item $TaskLoss_{n}$ is the task loss after incorporating the $n$-th fixation
point.
\end{itemize}

\subsection{Positional Encoding and Channel Encoding}

Each channel employs independent positional encoding. In the LRC,
positional encodings are assigned to the embeddings of image patches
based on their coordinates in the low-resolution image. In the HRC,
positional encodings are based on the relative coordinates of patches
within the Region of Interest (ROI).

When tokens from the LRC and HRC are transferred to the work memory
of the HE, a channel encoding is applied. For the LRC, all outputs
share a single channel encoding. In the HRC, the high-resolution image
is divided into smaller regions, with fixation points in different
regions treated as separate channels and assigned unique channel encodings.
Fixation points within the same region share the same channel encoding.

Both the positional encodings within channels and the channel encodings
are designed to be trainable.

\section{Experiments}

\subsection{Dataset and Parameter Settings}

The experiments were conducted on the MNIST dataset, which consists
of 10 classes, each containing 6,000 training images and 1,000 validation
images. The image dimensions were normalized to 32\texttimes 32 pixels.

The parameter settings are as follows:
\begin{itemize}
\item LRC Input: The original 32\texttimes 32 pixel image was downscaled
by a factor of 8, resulting in a 4\texttimes 4 low resolution global
image. The patch size for embedding was set to 1 pixel.
\item HRC Input: ROIs of 8\texttimes 8 pixels were cropped from the original
image. The patch size for embedding was set to 2 pixels.
\item Embedding Dimension: 32
\item Encoder Layers: The LRC, HRC, and HE each consist of 6 Transformer-based
encoder layers.
\item Number of Attention Heads: 4
\end{itemize}

\subsection{Effectiveness of the Fixation Mechanism}

\Tabref{Validation-Accuracy-of-MNIST} presents the classification
accuracy of TDFN on the MNIST validation set after training. The Fixation
Point Number indicates the number of fixation steps performed. When
the fixation point number is 0, only information from the Low-Resolution
Channel (LRC) is used for classification. Random Selected refers to
fixation points generated randomly, while FPG Generated refers to
fixation points generated by the trained Fixation Point Generator
(FPG). Coverage represents the proportion of the original image's
area covered by the fixation regions.

The following observations can be made:
\begin{enumerate}
\item Baseline Accuracy: When no fixation points are used (Fixation Point Number=0),
the model achieves a baseline accuracy of 68.03\%, relying solely
on low-resolution information from the LRC.
\item High-Resolution Information Importance: Incorporating high-resolution
information significantly boosts classification accuracy. For example,
even with two randomly selected fixation points, the accuracy improves
from 68.03\% to 74.00\%.
\item FPG Effectiveness: Fixation points generated by the trained FPG consistently
outperform randomly selected fixation points. For instance, with two
fixation points, accuracy increases from 74.00\% (random) to 84.20\%
(FPG-generated).
\item Performance Convergence Between FPG and Random Selection: As the number
of fixation points increases, the performance gap between FPG-generated
and randomly selected fixation points narrows. This occurs because
a larger number of fixation points allows random selection to cover
most of the image. For instance, at 16 fixation points, random selection
achieves an accuracy of 96.28\%, compared to 97.79\% for FPG-generated
points.
\end{enumerate}
\begin{table}
\centering
\caption{Validation Accuracy of MNIST on TDFN}\label{tab:Validation-Accuracy-of-MNIST}

\begin{tabular}{cccc}
\toprule 
Fixation Point Number & Random Selected & Generated by FPG & Coverage\tabularnewline
\midrule
\midrule 
0 & 68.03\% & 68.03\% & 0\%\tabularnewline
\midrule 
2 & 74.00\% & 84.20\% & 12.5\%\tabularnewline
\midrule 
4 & 86.41\% & 92.89\% & 25\%\tabularnewline
\midrule 
8 & 93.20\% & 96.11\% & 50\%\tabularnewline
\midrule 
12 & 95.29\% & 97.02\% & 75\%\tabularnewline
\midrule 
16 & 96.28\% & 97.79\% & 100\%\tabularnewline
\bottomrule
\end{tabular}
\end{table}

\subsection{Dynamic Termination of Fixation}

A mechanism for early termination of fixation can be introduced based
on the maximum classification probability (MCP) output by the classifier.
If this probability exceeds a predefined threshold (e.g., 0.9), the
fixation process can be terminated. This allows simpler samples to
require fewer fixation steps, while more complex samples may use additional
fixation steps. Such a mechanism significantly reduces computational
cost.

\Tabref{Average-Fixation-Steps} summarizes the average number of
fixation steps for different MCP threshold values. The results show
that achieving high recognition performance does not require high-resolution
analysis of the entire image for all samples. Instead, focusing on
small, high-resolution regions is sufficient, highlighting the efficiency
of the fixation mechanism. For example:
\begin{itemize}
\item At a MCP threshold of 0.9, the average number of fixation steps is
only 1.72, yielding a classification accuracy of 90.87\% and a coverage
rate of 10.76\%.
\item At a threshold of 0.98, the average number of fixation steps increases
to 4.51, achieving a classification accuracy of 97.20\% with a coverage
rate of 28.19\%.
\end{itemize}
\begin{table}
\centering

\caption{Average Fixation Steps, Classification Accuracy, and Coverage for
Different MCP Thresholds}\label{tab:Average-Fixation-Steps}

\begin{tabular}{cccc}
\toprule 
MCP Threshold & Average Fixation Steps & Classification Accuracy & Coverage\tabularnewline
\midrule
\midrule 
0.70 & 0.40 & 72.73\% & 2.48\%\tabularnewline
\midrule 
0.75 & 0.56 & 77.17\% & 3.50\%\tabularnewline
\midrule 
0.80 & 0.83 & 81.87\% & 5.21\%\tabularnewline
\midrule 
0.85 & 1.19 & 86.62\% & 7.42\%\tabularnewline
\midrule 
0.90 & 1.72 & 90.87\% & 10.76\%\tabularnewline
\midrule 
0.95 & 2.76 & 95.21\% & 17.22\%\tabularnewline
\midrule 
0.96 & 3.17 & 95.89\% & 19.78\%\tabularnewline
\midrule 
0.97 & 3.69 & 96.65\% & 23.04\%\tabularnewline
\midrule 
0.98 & 4.51 & 97.20\% & 28.19\%\tabularnewline
\midrule 
0.99 & 6.17 & 97.53\% & 38.54\%\tabularnewline
\bottomrule
\end{tabular}
\end{table}

\subsection{Visualization of Fixation Points}

\begin{figure}
\centering
\includegraphics[scale=0.6]{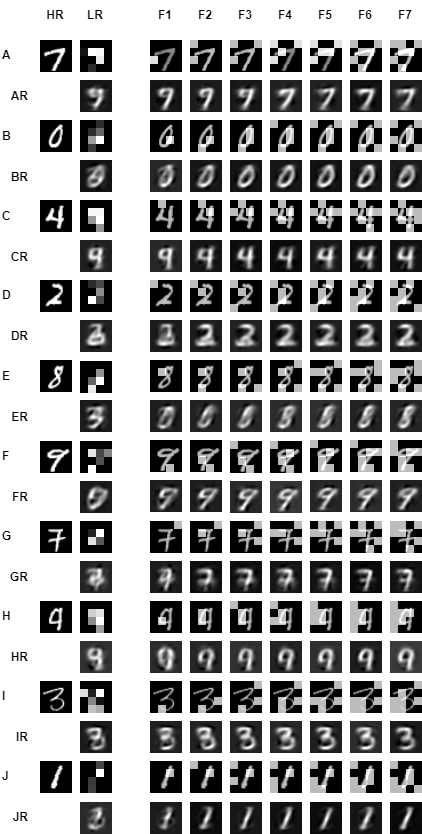}

\caption{Visualization of Fixation Points. The first column shows the original
input images. The second column displays the low-resolution images
(odd rows) and the reconstructed images generated by TDFN using only
the low-resolution inputs (even rows). The third to ninth columns
sequentially present the fixation points generated by the FPG (odd
rows, represented as light squares) and the reconstructed images generated
by TDFN using both low-resolution inputs and high-resolution inputs
from the fixation points (even rows). The reconstructed images are
displayed to illustrate how the addition of fixation points introduces
supplementary information.}\label{fig:Visualization-of-Fixation}
\end{figure}

\Figref{Visualization-of-Fixation} illustrates the fixation points
generated by the FPG and their contribution to image reconstruction.
It is evident that some fixation points are positioned on critical
regions such as the openings, endpoints, and external corners of characters,
while others fall on background areas. The reason behind this phenomenon
warrants further investigation in future work. A preliminary hypothesis
suggests that recognition relies not only on what should be present
but also on the absence of irrelevant or misleading features.

An additional noteworthy observation is that, at certain steps, the
inclusion of specific fixation points significantly enhances the quality
of the reconstructed image. This improvement reduces ambiguities in
the reconstruction, providing a more definitive indication of the
image’s category.

\section{Conclusion}

This study introduces the Task-Driven Fixation Network (TDFN) as an
efficient solution for complex tasks, leveraging a biologically inspired
fixation mechanism. Unlike traditional neural network architectures
that rely on exhaustive high-resolution analysis, TDFN selectively
incorporates high-resolution information from regions of interest,
guided by a Fixation Point Generator (FPG). This task-driven approach
balances computational efficiency with high task performance. Unlike
rule-based models or neural networks designed solely for generating
saliency maps or fixation points, TDFN effectively utilizes fixation
mechanisms for real classification tasks.

TDFN is built on standard Transformer modules with minimal modifications,
including the addition of a Fixation Point Generator, balancing the
training of classification and image reconstruction, and sequentially
appending high-resolution information through the High-Resolution
Channel (HRC). This design ensures that TDFN is both effective and
straightforward to implement.

Experimental results demonstrate the advantages of the fixation mechanism,
enabling TDFN to maintain high classification accuracy while significantly
reducing computational overhead. By leveraging low-resolution inputs
and a small number of high-resolution regions of interest, TDFN achieves
strong classification performance. The dynamic termination strategy
further enhances efficiency by tailoring fixation steps to the complexity
of individual samples.

The success of TDFN underscores the potential of fixation-inspired
mechanisms for scalable, task-specific neural networks. Future work
will focus on extending this framework to more challenging datasets
and applications, such as object detection and segmentation, while
exploring improved strategies for fixation point generation and dynamic
termination.

\bibliographystyle{plainnat}
\bibliography{TDFN}

\begin{thebibliography}{22}
\providecommand{\natexlab}[1]{#1}
\providecommand{\url}[1]{\texttt{#1}}
\expandafter\ifx\csname urlstyle\endcsname\relax
  \providecommand{\doi}[1]{doi: #1}\else
  \providecommand{\doi}{doi: \begingroup \urlstyle{rm}\Url}\fi

\bibitem[Ballard et~al.(1995)Ballard, Hayhoe, and
  Pelz]{ballardMemoryRepresentationsNatural1995}
Dana~H. Ballard, Mary~M. Hayhoe, and Jeff~B. Pelz.
\newblock Memory {{Representations}} in {{Natural Tasks}}.
\newblock \emph{Journal of Cognitive Neuroscience}, 7\penalty0 (1):\penalty0
  66--80, January 1995.

\bibitem[Dosovitskiy et~al.(2021)Dosovitskiy, Beyer, Kolesnikov, Weissenborn,
  Zhai, Unterthiner, Dehghani, Minderer, Heigold, Gelly, Uszkoreit, and
  Houlsby]{dosovitskiyImageWorth16x162021}
Alexey Dosovitskiy, Lucas Beyer, Alexander Kolesnikov, Dirk Weissenborn,
  Xiaohua Zhai, Thomas Unterthiner, Mostafa Dehghani, Matthias Minderer, Georg
  Heigold, Sylvain Gelly, Jakob Uszkoreit, and Neil Houlsby.
\newblock An {{Image}} is {{Worth}} 16x16 {{Words}}: {{Transformers}} for
  {{Image Recognition}} at {{Scale}}, June 2021.
\newblock arXiv:2010.11929.

\bibitem[Ehinger et~al.(2009)Ehinger, {Hidalgo-Sotelo}, Torralba, and
  Oliva]{ehingerModellingSearchPeople2009}
Krista~A. Ehinger, Barbara {Hidalgo-Sotelo}, Antonio Torralba, and Aude Oliva.
\newblock Modelling search for people in 900 scenes: {{A}} combined source
  model of eye guidance.
\newblock \emph{Visual Cognition}, 17\penalty0 (6-7):\penalty0 945--978, August
  2009.

\bibitem[Gao and
  Vasconcelos(2009)]{gaoDecisionTheoreticSaliencyComputational2009}
Dashan Gao and Nuno Vasconcelos.
\newblock Decision-{{Theoretic Saliency}}: {{Computational Principles}},
  {{Biological Plausibility}}, and {{Implications}} for {{Neurophysiology}} and
  {{Psychophysics}}.
\newblock \emph{Neural Computation}, 21\penalty0 (1):\penalty0 239--271,
  January 2009.

\bibitem[Huang et~al.(2015)Huang, Shen, Boix, and
  Zhao]{huangSALICONReducingSemantic2015}
Xun Huang, Chengyao Shen, Xavier Boix, and Qi~Zhao.
\newblock {{SALICON}}: {{Reducing}} the {{Semantic Gap}} in {{Saliency
  Prediction}} by {{Adapting Deep Neural Networks}}.
\newblock In \emph{2015 {{IEEE International Conference}} on {{Computer
  Vision}} ({{ICCV}})}, pages 262--270, Santiago, Chile, December 2015. IEEE.
\newblock ISBN 978-1-4673-8391-2.

\bibitem[Itti et~al.(1998)Itti, Koch, and
  Niebur]{ittiModelSaliencybasedVisual1998}
L.~Itti, C.~Koch, and E.~Niebur.
\newblock A model of saliency-based visual attention for rapid scene analysis.
\newblock \emph{IEEE Transactions on Pattern Analysis and Machine
  Intelligence}, 20\penalty0 (11):\penalty0 1254--1259, November 1998.

\bibitem[Koch and Ullman(1985)]{kochShiftsSelectiveVisual1985}
C.~Koch and S.~Ullman.
\newblock Shifts in selective visual attention: Towards the underlying neural
  circuitry.
\newblock \emph{Human Neurobiology}, 4\penalty0 (4):\penalty0 219--227, 1985.

\bibitem[K{\"u}mmerer et~al.(2015)K{\"u}mmerer, Theis, and
  Bethge]{kummererDeepGazeBoosting2015}
Matthias K{\"u}mmerer, Lucas Theis, and Matthias Bethge.
\newblock Deep {{Gaze I}}: {{Boosting Saliency Prediction}} with {{Feature Maps
  Trained}} on {{ImageNet}}, April 2015.
\newblock arXiv:1411.1045.

\bibitem[K{\"u}mmerer et~al.(2016)K{\"u}mmerer, Wallis, and
  Bethge]{kummererDeepGazeIIReading2016}
Matthias K{\"u}mmerer, Thomas S.~A. Wallis, and Matthias Bethge.
\newblock {{DeepGaze II}}: {{Reading}} fixations from deep features trained on
  object recognition, October 2016.
\newblock arXiv.1610.01563.

\bibitem[Oliva et~al.(2003)Oliva, Torralba, Castelhano, and
  Henderson]{olivaTopdownControlVisual2003}
A.~Oliva, A.~Torralba, M.S. Castelhano, and J.M. Henderson.
\newblock Top-down control of visual attention in object detection.
\newblock In \emph{Proceedings 2003 {{International Conference}} on {{Image
  Processing}} ({{Cat}}. {{No}}.{{03CH37429}})}, volume~1, pages I--253--6,
  Barcelona, Spain, 2003. IEEE.
\newblock ISBN 978-0-7803-7750-9.

\bibitem[Pan et~al.(2018)Pan, Ferrer, McGuinness, O'Connor, Torres, Sayrol, and
  {Giro-i-Nieto}]{panSalGANVisualSaliency2018}
Junting Pan, Cristian~Canton Ferrer, Kevin McGuinness, Noel~E. O'Connor, Jordi
  Torres, Elisa Sayrol, and Xavier {Giro-i-Nieto}.
\newblock {{SalGAN}}: {{Visual Saliency Prediction}} with {{Generative
  Adversarial Networks}}, July 2018.
\newblock arXiv:1701.01081.

\bibitem[Peters and Itti(2007)]{petersBottomupIncorporatingTaskdependent2007}
Robert~J. Peters and Laurent Itti.
\newblock Beyond bottom-up: {{Incorporating}} task-dependent influences into a
  computational model of spatial attention.
\newblock In \emph{2007 {{IEEE Conference}} on {{Computer Vision}} and
  {{Pattern Recognition}}}, pages 1--8, Minneapolis, MN, USA, June 2007. IEEE.
\newblock ISBN 978-1-4244-1179-5.

\bibitem[Rayner(1998)]{raynerEyeMovementsReading1998}
Keith Rayner.
\newblock Eye movements in reading and information processing: 20 years of
  research.
\newblock \emph{Psychological Bulletin}, 124\penalty0 (3):\penalty0 372--422,
  1998.

\bibitem[Treisman and
  Gelade(1980)]{treismanFeatureintegrationTheoryAttention1980}
Anne~M. Treisman and Garry Gelade.
\newblock A feature-integration theory of attention.
\newblock \emph{Cognitive Psychology}, 12\penalty0 (1):\penalty0 97--136,
  January 1980.

\bibitem[Vaswani et~al.(2017)Vaswani, Shazeer, Parmar, Uszkoreit, Jones, Gomez,
  Kaiser, and Polosukhin]{vaswaniAttentionAllYou2017}
Ashish Vaswani, Noam Shazeer, Niki Parmar, Jakob Uszkoreit, Llion Jones,
  Aidan~N. Gomez, Lukasz Kaiser, and Illia Polosukhin.
\newblock Attention {{Is All You Need}}, June 2017.
\newblock arXiv:1706.03762.

\bibitem[Vig et~al.(2014)Vig, Dorr, and
  Cox]{vigLargeScaleOptimizationHierarchical2014}
Eleonora Vig, Michael Dorr, and David Cox.
\newblock Large-{{Scale Optimization}} of {{Hierarchical Features}} for
  {{Saliency Prediction}} in {{Natural Images}}.
\newblock In \emph{2014 {{IEEE Conference}} on {{Computer Vision}} and
  {{Pattern Recognition}}}, pages 2798--2805, Columbus, OH, USA, June 2014.
  IEEE.
\newblock ISBN 978-1-4799-5118-5.

\bibitem[Wang et~al.(2020)Wang, Shen, Dong, Borji, and
  Yang]{wangInferringSalientObjects2020}
Wenguan Wang, Jianbing Shen, Xingping Dong, Ali Borji, and Ruigang Yang.
\newblock Inferring {{Salient Objects}} from {{Human Fixations}}.
\newblock \emph{IEEE Transactions on Pattern Analysis and Machine
  Intelligence}, 42\penalty0 (8):\penalty0 1913--1927, August 2020.

\bibitem[Wang et~al.(2021)Wang, Shen, Xie, Cheng, Ling, and
  Borji]{wangRevisitingVideoSaliency2021}
Wenguan Wang, Jianbing Shen, Jianwen Xie, Ming-Ming Cheng, Haibin Ling, and Ali
  Borji.
\newblock Revisiting {{Video Saliency Prediction}} in the {{Deep Learning
  Era}}.
\newblock \emph{IEEE Transactions on Pattern Analysis and Machine
  Intelligence}, 43\penalty0 (1):\penalty0 220--237, January 2021.

\bibitem[Yan et~al.(2021)Yan, Chen, Xiao, Qi, Wang, and
  Xiao]{yanReviewVisualSaliency2021}
Fei Yan, Cheng Chen, Peng Xiao, Siyu Qi, Zhiliang Wang, and Ruoxiu Xiao.
\newblock Review of {{Visual Saliency Prediction}}: {{Development Process}}
  from {{Neurobiological Basis}} to {{Deep Models}}.
\newblock \emph{Applied Sciences}, 12\penalty0 (1):\penalty0 309, December
  2021.

\bibitem[Yarbus(1967)]{yarbusEyeMovementsVision1967}
A.~L. Yarbus.
\newblock \emph{Eye {{Movements}} and {{Vision}}}.
\newblock Springer, New York, NY, 1967.
\newblock ISBN 978-1-4899-5379-7.

\bibitem[{Yulin Xie} et~al.(2013){Yulin Xie}, {Huchuan Lu}, and {Ming-Hsuan
  Yang}]{yulinxieBayesianSaliencyLow2013}
{Yulin Xie}, {Huchuan Lu}, and {Ming-Hsuan Yang}.
\newblock Bayesian {{Saliency}} via {{Low}} and {{Mid Level Cues}}.
\newblock \emph{IEEE Transactions on Image Processing}, 22\penalty0
  (5):\penalty0 1689--1698, May 2013.

\bibitem[Zhang et~al.(2008)Zhang, Tong, Marks, Shan, and
  Cottrell]{zhangSUNBayesianFramework2008}
Lingyun Zhang, Matthew~H. Tong, Tim~K. Marks, Honghao Shan, and Garrison~W.
  Cottrell.
\newblock {{SUN}}: {{A Bayesian}} framework for saliency using natural
  statistics.
\newblock \emph{Journal of Vision}, 8\penalty0 (7):\penalty0 32, December 2008.

\end{thebibliography}

\end{document}